\documentclass[11pt]{article}
\usepackage{eamt20}
\usepackage{times}
\usepackage{url}
\usepackage{latexsym}
\usepackage[small,bf]{caption} % added MLF 20171211
\usepackage{booktabs}
\usepackage{multirow}
\usepackage{subcaption}
\usepackage{graphicx}
\usepackage{overpic}
\usepackage[T2A,T1]{fontenc}
\usepackage[utf8]{inputenc}
\usepackage[russian,english]{babel}
\usepackage{ulem}
\usepackage{CJKutf8}

\title{When and Why is Unsupervised Neural Machine Translation Useless?}

\author{Yunsu Kim \hspace{9pt} Miguel Gra\c{c}a$^\dagger$ \hspace{5pt} Hermann Ney\\
  Human Language Technology and Pattern Recognition Group \\
  RWTH Aachen University, Aachen, Germany \\
  {\tt \{surname\}@cs.rwth-aachen.de}}

\date{}

\begin{document}
\maketitle
\begin{abstract}
  This paper studies the practicality of the current state-of-the-art unsupervised methods in neural machine translation (NMT).
  In ten translation tasks with various data settings, we analyze the conditions under which the unsupervised methods fail to produce reasonable translations.
  We show that their performance is severely affected by linguistic dissimilarity and domain mismatch between source and target monolingual data.
  Such conditions are common for low-resource language pairs, where unsupervised learning works poorly.
  In all of our experiments, supervised and semi-supervised baselines with 50k-sentence bilingual data outperform the best unsupervised results.
  Our analyses pinpoint the limits of the current unsupervised NMT and also suggest immediate research directions.
\end{abstract}

\section{Introduction}

Statistical methods for machine translation (MT) require a large set of sentence pairs in two languages to build a decent translation system \cite{resnik2003web,koehn2005europarl}.
Such bilingual data is scarce for most language pairs and its quality varies largely over different domains \cite{al-onaizan2002translation,chu2018survey}.
Neural machine translation (NMT) \cite{bahdanau2015neural,vaswani2017attention}, the standard paradigm of MT these days, has been claimed to suffer from the data scarcity more severely than phrase-based MT \cite{koehn2017six}.

Unsupervised NMT, which trains a neural translation model only with monolingual corpora, was proposed for those scenarios which lack bilingual data \cite{artetxe2018unsupervised-nmt,lample2018unsupervised}.
Despite its progress in research, the performance of the unsupervised methods has been evaluated mostly on high-resource language pairs, e.g. German$\leftrightarrow$English or French$\leftrightarrow$English \cite{artetxe2018unsupervised-nmt,lample2018unsupervised,yang2018unsupervised,artetxe2018unsupervised-smt,lample2018phrasebased,ren2019unsupervised,artetxe2019effective,sun2019unsupervised,sen2019multilingual}.
For these language pairs, huge bilingual corpora are already available, so there is no need for unsupervised learning in practice.
Empirical results in these tasks do not carry over to low-resource language pairs; they simply fail to produce any meaningful translations \cite{neubig2018rapid,guzman2019flores}.

This paper aims for a more comprehensive and pragmatic study on the performance of unsupervised NMT.
Our experiments span ten translation tasks in the following five language pairs:
\begin{itemize}\itemsep0em
  \item German$\leftrightarrow$English: similar languages, abundant bilingual/monolingual data
  \item Russian$\leftrightarrow$English: distant languages, abundant bilingual/monolingual data, similar sizes of the alphabet
  \item Chinese$\leftrightarrow$English: distant languages, abundant bilingual/monolingual data, very different sizes of the alphabet
  \item Kazakh$\leftrightarrow$English: distant languages, scarce bilingual data, abundant monolingual data
  \item Gujarati$\leftrightarrow$English: distant languages, scarce bilingual/monolingual data
\end{itemize}
For each task, we compare the unsupervised performance with its supervised and semi-supervised counterparts.
In addition, we make the monolingual training data vary in size and domain to cover many more scenarios, showing under which conditions unsupervised NMT works poorly.

Here is a summary of our contributions:\vspace{-0.1em}
\begin{itemize}\itemsep0em
  \item We thoroughly evaluate the performance of state-of-the-art unsupervised NMT in numerous real and artificial translation tasks.
  \item We provide guidelines on whether to employ unsupervised NMT in practice, by showing how much bilingual data is sufficient to outperform the unsupervised results.
  \item We clarify which factors make unsupervised NMT weak and which points must be improved, by analyzing the results both quantitatively and qualitatively.
\end{itemize}

\section{Related Work}
\label{sec:related}

The idea of unsupervised MT dates back to word-based decipherment methods \cite{knight2006unsupervised,ravi2011deciphering}.
They learn only lexicon models at first, but add alignment models \cite{dou2014beyond,nuhn2019unsupervised} or heuristic features \cite{naim2018feature} later.
Finally, \newcite{artetxe2018unsupervised-smt} and \newcite{lample2018phrasebased} train a fully-fledged phrase-based MT system in an unsupervised way.

With neural networks, unsupervised learning of a sequence-to-sequence NMT model has been proposed by \newcite{lample2018unsupervised} and \newcite{artetxe2018unsupervised-nmt}.
Though having slight variations \cite{yang2018unsupervised,sun2019unsupervised,sen2019multilingual}, unsupervised NMT approaches commonly 1) learn a shared model for both source$\rightarrow$target and target$\rightarrow$source 2) using iterative back-translation, along with 3) a denoising autoencoder objective.
They are initialized with either cross-lingual word embeddings or a cross-lingual language model (LM).
To further improve the performance at the cost of efficiency, \newcite{lample2018phrasebased}, \newcite{ren2019unsupervised} and \newcite{artetxe2019effective} combine unsupervised NMT with unsupervised phrase-based MT.
On the other hand, one can also avoid the long iterative training by applying a separate denoiser directly to the word-by-word translations from cross-lingual word embeddings \cite{kim2018improving,pourdamghani2019translating}.

Unsupervised NMT approaches have been so far evaluated mostly on high-resource language pairs, e.g. French$\rightarrow$English, for academic purposes.
In terms of practicality, they tend to underperform in low-resource language pairs, e.g. Azerbaijani$\rightarrow$English \cite{neubig2018rapid} or Nepali$\rightarrow$English \cite{guzman2019flores}.
To the best of our knowledge, this work is the first to systematically evaluate and analyze unsupervised learning for NMT in various data settings.

\section{Unsupervised NMT}
\label{sec:unmt}

This section reviews the core concepts of the recent unsupervised NMT framework and describes to which points they are potentially vulnerable.

\subsection{Bidirectional Modeling}
\label{sec:unmt:bidir}

Most of the unsupervised NMT methods share the model parameters between source$\rightarrow$target and target$\rightarrow$source directions.
They also often share a joint subword vocabulary across the two languages \cite{sennrich2016neural}.

Sharing a model among different translation tasks has been shown to be effective in multilingual NMT \cite{firat2016multi,johnson2017google,aharoni2019massively}, especially in improving performance on low-resource language pairs.
This is due to the commonality of natural languages; learning to represent a language is helpful to represent other languages, e.g. by transferring knowledge of general sentence structures.
It also provides good regularization for the model.

Unsupervised learning is an extreme scenario of MT, where bilingual information is very weak.
To supplement the weak and noisy training signal, knowledge transfer and regularization are crucial, which can be achieved by the bidirectional sharing.
It is based on the fact that a translation problem is dual in nature; source$\rightarrow$target and target$\rightarrow$source tasks are conceptually related to each other.

Previous works on unsupervised NMT vary in the degree of sharing: the whole encoder \cite{artetxe2018unsupervised-nmt,sen2019multilingual}, the middle layers \cite{yang2018unsupervised,sun2019unsupervised}, or the whole model \cite{lample2018unsupervised,lample2018phrasebased,ren2019explicit,conneau2019crosslingual}.

Note that the network sharing is less effective among linguistically distinct languages in NMT \cite{kocmi2018trivial,kim2019effective}.
It still works as a regularizer, but transferring knowledge is harder if the morphology or word order is quite different.
We show how well unsupervised NMT performs on such language pairs in Section \ref{sec:exp:semi}.

\subsection{Iterative Back-Translation}
\label{sec:unmt:bt}

Unsupervised learning for MT assumes no bilingual data for training.
A traditional remedy for the data scarcity is generating synthetic bilingual data from monolingual text \cite{koehn2005europarl,schwenk2008investigations,sennrich2016improving}.
To train a bidirectional model of Section \ref{sec:unmt:bidir}, we need bilingual data of both translation directions.
Therefore, most unsupervised NMT methods back-translate in both directions, i.e. source and target monolingual data to target and source language, respectively.

\begin{figure}[!ht]
    \centering
    \begin{subfigure}[b]{0.49\linewidth}
        \centering
        \begin{overpic}[width=0.7\linewidth]{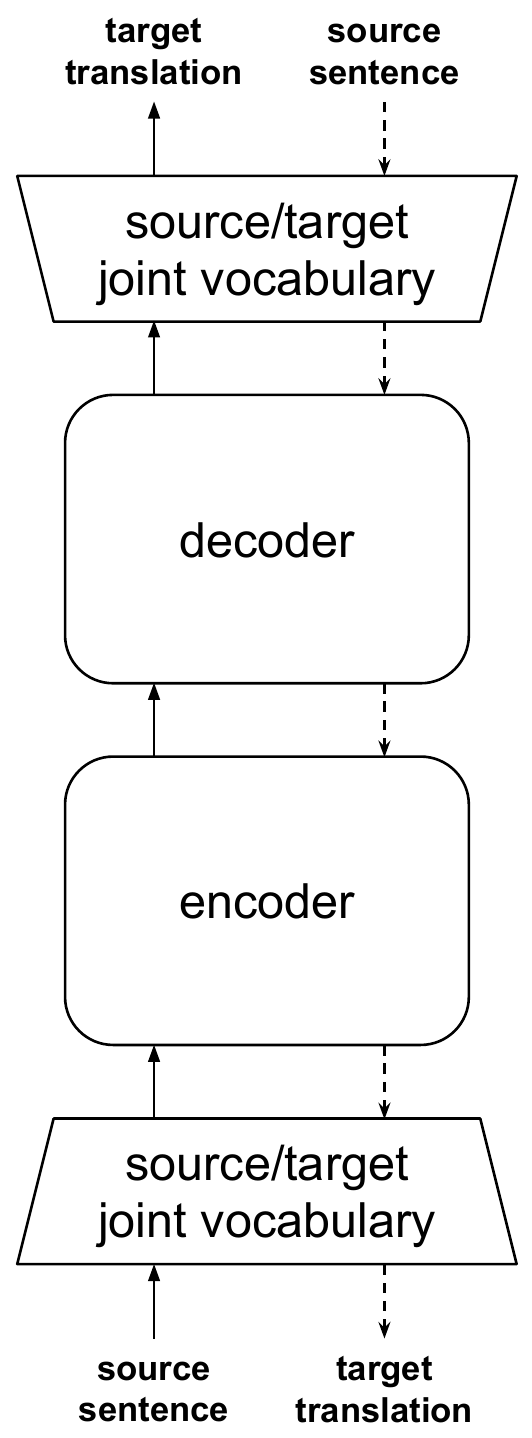}
        \put (2, 49) {\scriptsize 1)}
        \put (32, 49) {\scriptsize 2)}
        \end{overpic}
        \caption{}
    \end{subfigure}
    \begin{subfigure}[b]{0.49\linewidth}
        \centering
        \begin{overpic}[width=0.7\linewidth]{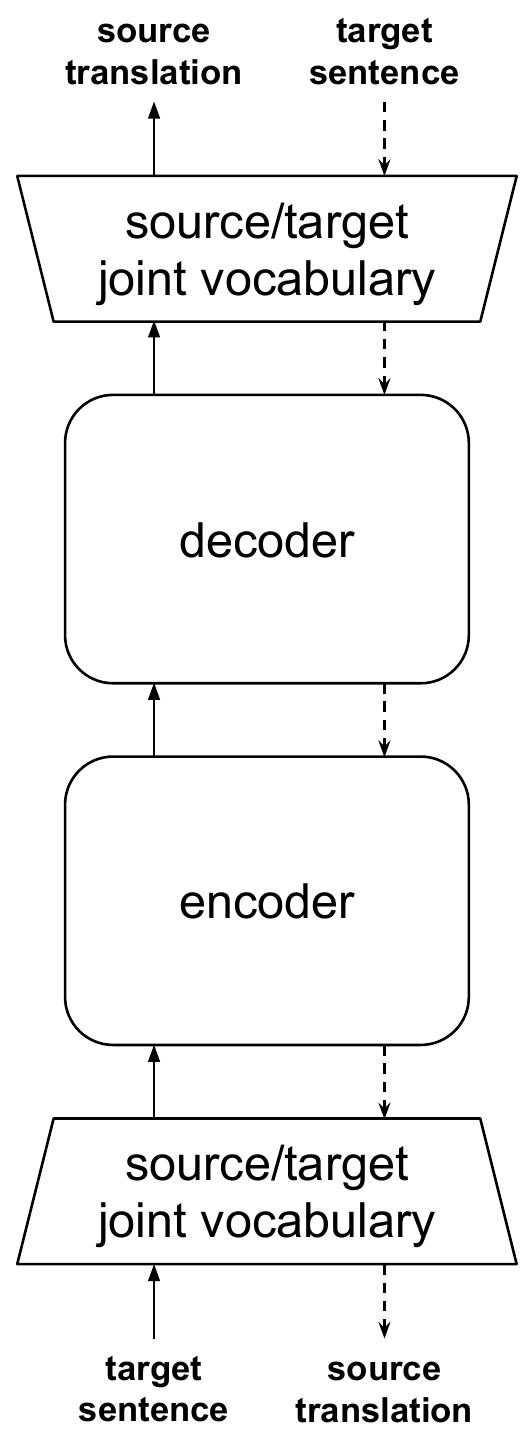}
        \put (2, 49) {\scriptsize 1)}
        \put (32, 49) {\scriptsize 2)}
        \end{overpic}
        \caption{}
    \end{subfigure}
    \caption{Iterative back-translation for training a bidirectional sequence-to-sequence model. The model first translates monolingual sentences (solid arrows), and then gets trained with the translation as the input and the original as the output (dashed arrows). This procedure alternates between \textbf{(a)} source$\rightarrow$target and \textbf{(b)} target$\rightarrow$source translations.}
    \label{fig:bt}
\end{figure}

In unsupervised learning, the synthetic data should be created not only once at the beginning but also repeatedly throughout the training.
At the early stages of training, the model might be too weak to generate good translations.
Hence, most methods update the training data as the model gets improved during training.
The improved model for source$\rightarrow$target direction back-translates source monolingual data, which improves the model for target$\rightarrow$source direction, and vice versa.
This cycle is called dual learning \cite{he2016dual} or iterative back-translation \cite{hoang2018iterative}.
Figure \ref{fig:bt} shows the case when it is applied to a fully shared bidirectional model.

One can tune the amount of back-translations per iteration: a mini-batch \cite{artetxe2018unsupervised-nmt,yang2018unsupervised,conneau2019crosslingual,ren2019explicit}, the whole monolingual data \cite{lample2018unsupervised,lample2018phrasebased,sun2019unsupervised}, or some size in between \cite{artetxe2019effective,ren2019unsupervised}.

However, even if carefully scheduled, the iterative training cannot recover from a bad optimum if the initial model is too poor.
Experiments in Section \ref{sec:exp:init-bt} highlight such cases.

\subsection{Initialization}
\label{sec:unmt:init}

To kickstart the iterative training, the model should be able to generate meaningful translations already in the first iteration.
We cannot expect the training to progress from a randomly initialized network and the synthetic data generated by it.

Cross-lingual embeddings give a good starting point for the model by defining a joint continuous space shared by multiple languages.
Ideally, in such a space, close embedding vectors are semantically related to each other regardless of their languages; they can be possible candidates for translation pairs \cite{mikolov2013efficient}.
It can be learned either in word level \cite{artetxe2017learning,conneau2018word} or in sentence level \cite{conneau2019crosslingual} using only monolingual corpora.

In the word level, we can initialize the embedding layers with cross-lingual word embedding vectors \cite{artetxe2018unsupervised-nmt,lample2018unsupervised,yang2018unsupervised,lample2018phrasebased,artetxe2019effective,sun2019unsupervised}.
On the other hand, the whole encoder/decoder parameters can be initialized with cross-lingual sequence training \cite{conneau2019crosslingual,ren2019explicit,song2019mass}.

Cross-lingual word embedding has limited performance among distant languages \cite{sogaard2018limitations,nakashole2018characterizing} and so does cross-lingual LM \cite{pires2019multilingual}.
Section \ref{sec:exp:init-bt} shows the impact of a poor initialization.

\begin{table*}[!ht]
  \centering
  \fontsize{7.8}{10}\selectfont 
  \setlength\tabcolsep{4pt}
  \begin{tabular}{cccccccccccc}
  \toprule
    & & \multicolumn{2}{c}{\bf de-en} & \multicolumn{2}{c}{\bf ru-en} & \multicolumn{2}{c}{\bf zh-en} & \multicolumn{2}{c}{\bf kk-en} & \multicolumn{2}{c}{\bf gu-en}\\
    \cmidrule(l{2pt}r{2pt}){3-4}\cmidrule(l{2pt}r{2pt}){5-6}\cmidrule(l{2pt}r{2pt}){7-8}\cmidrule(l{2pt}r{2pt}){9-10}\cmidrule(l{2pt}r{2pt}){11-12}
    & & German & English & Russian & English & Chinese & English & Kazakh & English & Gujarati & English\\
    \midrule
    \multicolumn{2}{c}{Language family} & Germanic & Germanic & Slavic & Germanic & Sinitic & Germanic & Turkic & Germanic & Indic & Germanic\\
    \multicolumn{2}{c}{Alphabet Size} & 60 & 52 & 66 & 52 & 8,105 & 52 & 42 & 52 & 91 & 52\\
    \midrule
    \multirow{2}{*}{Monolingual} & Sentences & \multicolumn{2}{c}{100M} & \multicolumn{2}{c}{71.6M} & \multicolumn{2}{c}{30.8M} & \multicolumn{2}{c}{18.5M} & \multicolumn{2}{c}{4.1M}\\
    & Words & 1.8B & 2.3B & 1.1B & 2.0B & 1.4B & 699M & 278.5M & 421.5M & 121.5M & 93.8M\\
    \midrule
    \multirow{2}{*}{Bilingual} & Sentences & \multicolumn{2}{c}{5.9M} & \multicolumn{2}{c}{25.4M} & \multicolumn{2}{c}{18.9M} & \multicolumn{2}{c}{222k} & \multicolumn{2}{c}{156k}\\
    & Words & 137.4M & 144.9M & 618.6M & 790M & 440.3M & 482.9M & 1.6M & 1.9M & 2.3M & 1.5M\\
    \bottomrule
  \end{tabular}
\caption{Training data statistics.}
\label{tab:corpus}
\vspace{-0.5em}
\end{table*}

\subsection{Denoising Autoencoder}
\label{sec:unmt:dae}

Initializing the word embedding layers furnishes the model with cross-lingual matching in the lexical embedding space, but does not provide any information on word orders or generation of text.
Cross-lingual LMs encode word sequences in different languages, but they are not explicitly trained to reorder source words to the target language syntax.
Both ways do not initialize the crucial parameters for reordering: the encoder-decoder attention and the recurrence on decoder states.

As a result, an initial model for unsupervised NMT tends to generate word-by-word translations with little reordering, which are very non-fluent when source and target languages have distinct word orders.
Training on such data discourages the model from reordering words, which might cause a vicious cycle by generating even less-reordered synthetic sentence pairs in the next iterations. 

\begin{figure}[!ht]
    \centering
    \includegraphics[width=0.35\linewidth]{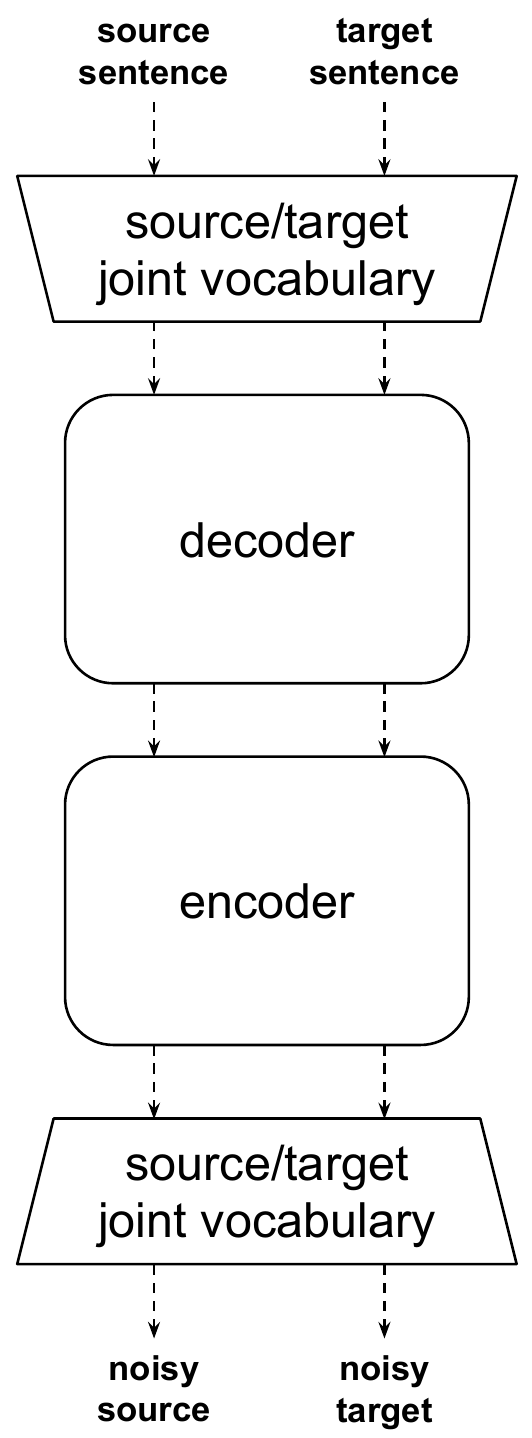}
    \caption{Denoising autoencoder training for source or target language.}
    \label{fig:dae}
    \vspace{-0.3em}
\end{figure}

Accordingly, unsupervised NMT employs an additional training objective of denoising autoencoding \cite{hill2016learning}.
Given a clean sentence, artificial noises are injected, e.g. deletion or permutation of words, to make a corrupted input.
The denoising objective trains the model to reorder the noisy input to the correct syntax, which is essential for generating fluent outputs.
This is done for each language individually with monolingual data, as shown in Figure \ref{fig:dae}.

Once the model is sufficiently trained for denoising, it is helpful to remove the objective or reduce its weight \cite{graca2018rwth}.
At the later stages of training, the model gets improved in reordering and translates better; learning to denoise might hurt the performance in clean test sets.

\begin{table*}[!ht]
  \centering
  \setlength\tabcolsep{5pt}
  \begin{tabular}{lccccccccccc}
    \toprule
    & & \multicolumn{10}{c}{\textsc{Bleu} [\%]}\\
    \cmidrule{3-12}
    Approach & & {\bf de-en} & {\bf en-de} & {\bf ru-en} & {\bf en-ru} & {\bf zh-en} & {\bf en-zh} & {\bf kk-en} & {\bf en-kk} & {\bf gu-en} & {\bf en-gu}\\
    \midrule
    Supervised & & 39.5 & 39.1 & 29.1 & 24.7 & 26.2 & 39.6 & 10.3 & 2.4 & 9.9 & 3.5\\
    Semi-supervised & & 43.6 & 41.0 & 30.8 & 28.8 & 25.9 & 42.7 & 12.5 & 3.1 & 14.2 & 4.0\\
    Unsupervised & & 23.8 & 20.2 & 12.0 & 9.4 & 1.5 & 2.5 & 2.0 & 0.8 & 0.6 & 0.6\\
    \bottomrule
  \end{tabular}
\caption{Comparison among supervised, semi-supervised, and unsupervised learning. All bilingual data was used for the (semi-)supervised results and all monolingual data was used for the unsupervised results (see Table \ref{tab:corpus}). All results are computed on newstest2019 of each task, except for de-en/en-de and ru-en/en-ru on newstest2018.}%ch-bleu for en-zh?
\label{tab:overview}
\vspace{-0.5em}
\end{table*}

\section{Experiments and Analysis}
\label{sec:exp}

\noindent\textbf{Data}\hspace{0.5cm} Our experiments were conducted on WMT 2018 German$\leftrightarrow$English and Russian$\leftrightarrow$ English, WMT 2019 Chinese$\leftrightarrow$English, Kazakh$\leftrightarrow$ English, and Gujarati$\leftrightarrow$English (Table \ref{tab:corpus}).
We preprocessed the data using the \textsc{Moses}\footnote{http://www.statmt.org/moses} tokenizer and a frequent caser.
For Chinese, we used the \textsc{jieba} segmenter\footnote{https://github.com/fxsjy/jieba}.
Lastly, byte pair encoding (BPE) \cite{sennrich2016neural} was learned jointly over source and target languages with 32k merges and applied without vocabulary threshold.
\vspace{0.7em}

\noindent\textbf{Model}\hspace{0.5cm} We used 6-layer Transformer base architecture \cite{vaswani2017attention} by default: 512-dimension embedding/hidden layers, 2048-dimension feedforward sublayers, and 8 heads.
\vspace{0.7em}

\noindent\textbf{Decoding and Evaluation}\hspace{0.5cm} Decoding was done with beam size 5.
We evaluated the test performance with \textsc{SacreBleu} \cite{post2018call}.
\vspace{0.7em}

\noindent\textbf{Unsupervised Learning}\hspace{0.5cm} We ran \textsc{Xlm}\footnote{https://github.com/facebookresearch/XLM} by \newcite{conneau2019crosslingual} for the unsupervised experiments.
The back-translations were done with beam search for each mini-batch of 16k tokens.
The weight of the denoising objective started with 1 and linearly decreased to 0.1 until 100k updates, and then decreased to 0 until 300k updates.

The model's encoder and decoder were both initialized with the same pre-trained cross-lingual LM.
We removed the language embeddings from the encoder for better cross-linguality (see Section \ref{sec:exp:examples}).
Unless otherwise specified, we used the same monolingual training data for both pre-training and translation training.
For the pre-training, we set the batch size to 256 sentences (around 66k tokens).

Training was done with Adam \cite{kingma2014adam} with an initial learning rate of 0.0001, where dropout \cite{srivastava2014dropout} of probability 0.1 was applied to each layer output and attention components.
With a checkpoint frequency of 200k sentences, we stopped the training when the validation perplexity (pre-training) or \textsc{Bleu} (translation training) was not improved for ten checkpoints.
We extensively tuned the hyperparameters for a single GPU with 12GB memory, which is widely applicable to moderate industrial/academic environments.
All other hyperparameter values follow the recommended settings of \textsc{Xlm}.
\vspace{0.7em}

\noindent\textbf{Supervised Learning}\hspace{0.5cm} Supervised experiments used the same hyperparameters as the unsupervised learning, except 12k tokens for the batch size, 0.0002 for the initial learning rate, and 10k batches for each checkpoint.

If the bilingual training data contains less than 500k sentence pairs, we reduced the BPE merges to 8k, the batch size to 2k, and the checkpoint frequency to 4k batches; we also increased the dropout rate to 0.3 \cite{sennrich2019revisiting}.

\vspace{0.7em}
\noindent\textbf{Semi-supervised Learning}\hspace{0.5cm} Semi-supervised experiments continued the training from the supervised baseline with back-translations added to the training data.
We used 4M back-translated sentences for the low-resource cases, i.e. if the original bilingual data has less than 500k lines, and 10M back-translated sentences otherwise.

\subsection{Unsupervised vs. (Semi-)Supervised}
\label{sec:exp:semi}

We first address the most general question of this paper: For NMT, can unsupervised learning replace semi-supervised or supervised learning?
Table \ref{tab:overview} compares the unsupervised performance to simple supervised and semi-supervised baselines.

In all tasks, unsupervised learning shows much worse performance than (semi-)supervised learning.
It produces readable translations in two high-resource language pairs (German$\leftrightarrow$English and Russian$\leftrightarrow$English), but their scores are only around half of the semi-supervised systems.
In other three language pairs, unsupervised NMT fails to converge at any meaningful optimum, reaching less than 3\% \textsc{Bleu} scores.
Note that, in these three tasks, source and target languages are very different in the alphabet, morphology, and word order, etc.
The results in Kazakh$\leftrightarrow$English and Gujarati$\leftrightarrow$English show that the current unsupervised NMT cannot be an alternative to (semi-)supervised NMT in low-resource conditions.

\begin{figure}[!t]
    \centering
    \begin{subfigure}[t]{0.4\textwidth}
    \hspace{-0.3cm}
    \includegraphics[width=\textwidth]{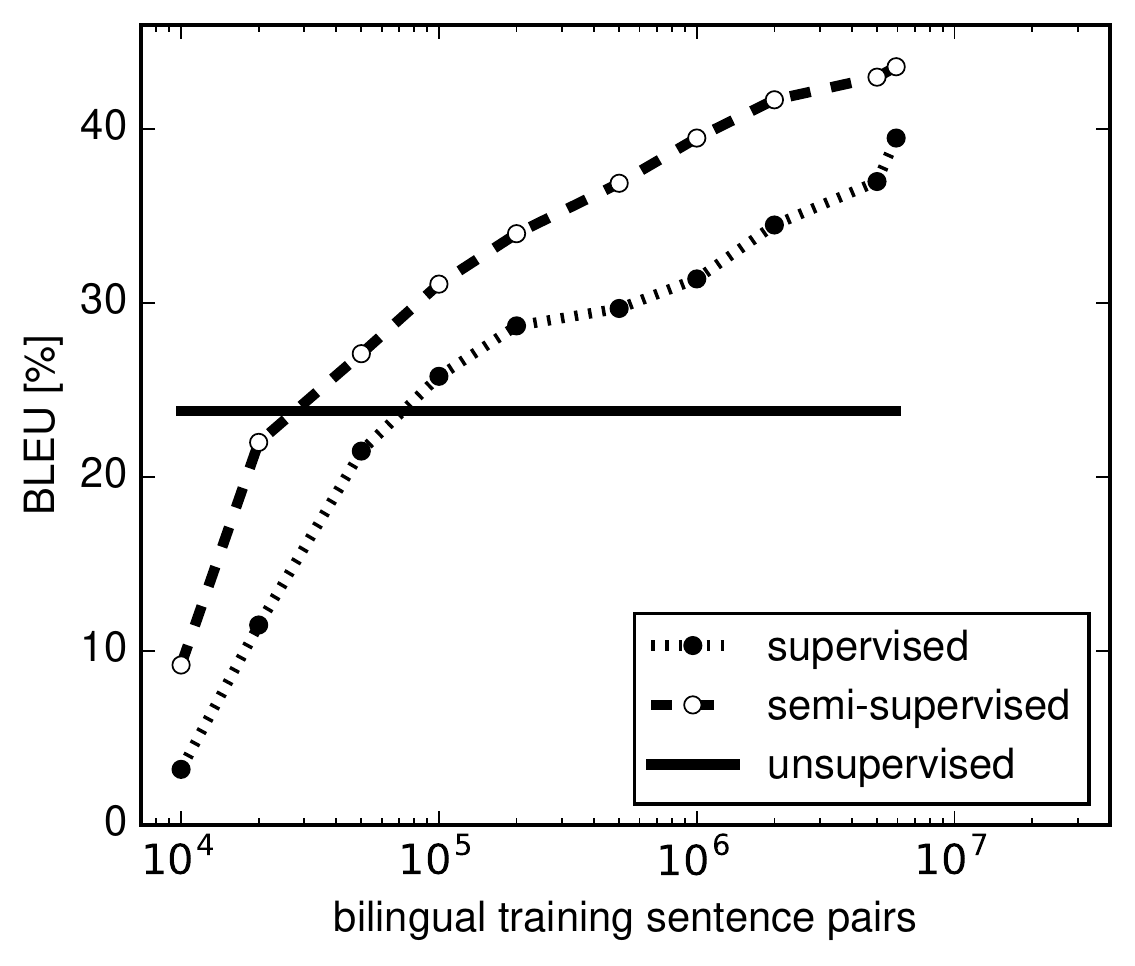}
    \caption{\label{fig:unsup-sup-semisup:de}German$\rightarrow$English}\vspace{0.5em}
    \end{subfigure}
    \begin{subfigure}[t]{0.4\textwidth}
    \hspace{-0.3cm}
    \includegraphics[width=\textwidth]{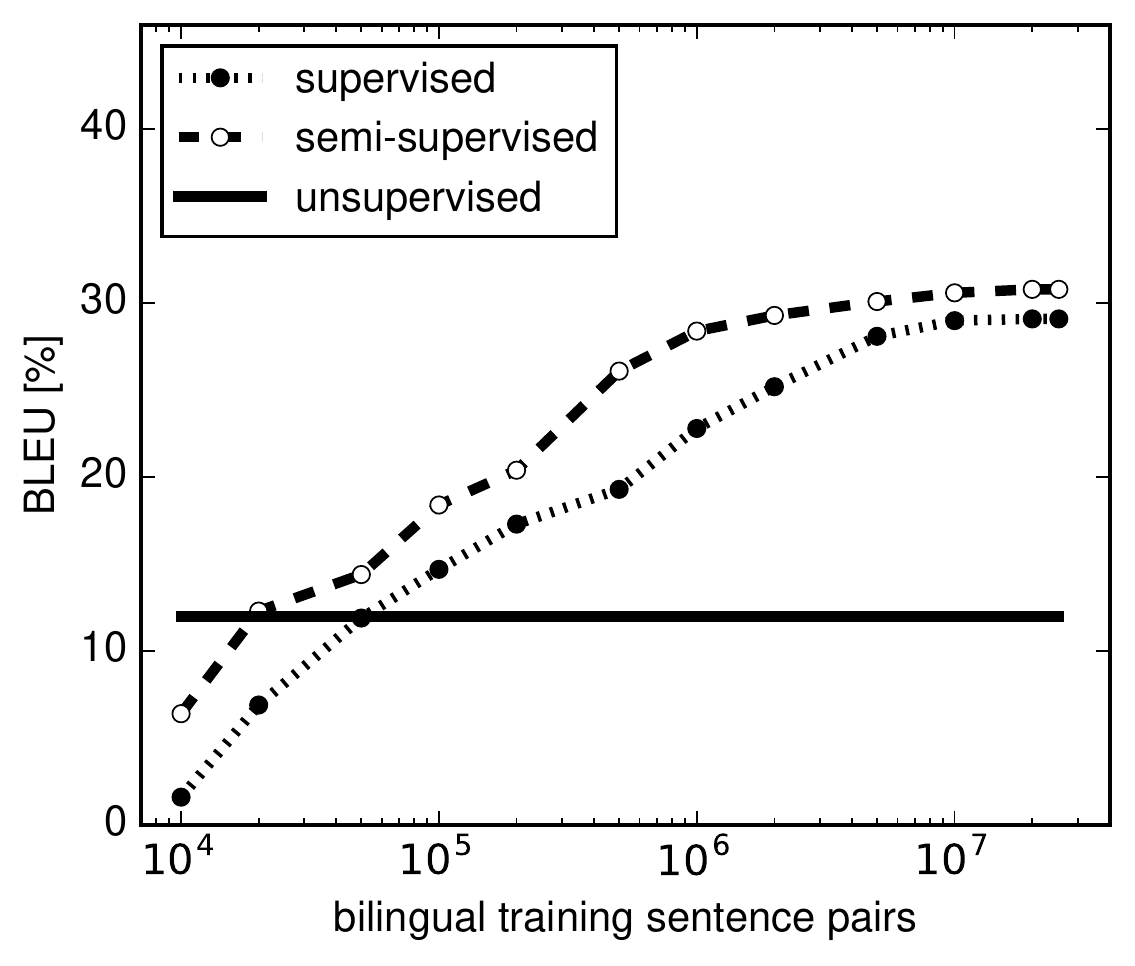}
    \caption{\label{fig:unsup-sup-semisup:ru}Russian$\rightarrow$English}
    \end{subfigure}
    \caption{Supervised and semi-supervised learning over bilingual training data size. Unsupervised learning (horizontal line) uses all monolingual data of Table \ref{tab:corpus}.}
    \label{fig:unsup-sup-semisup}
    \vspace{-1em}
\end{figure}

To discover the precise condition where the unsupervised learning is useful in practice, we vary the size of the given bilingual training data for (semi-)supervised learning and plot the results in Figure \ref{fig:unsup-sup-semisup}.
Once we have 50k bilingual sentence pairs in German$\leftrightarrow$English, simple semi-supervised learning already outperforms unsupervised learning with 100M monolingual sentences in each language.
Even without back-translations (supervised), 100k-sentence bilingual data is sufficient to surpass unsupervised NMT.

In the Russian$\leftrightarrow$English task, the unsupervised learning performance can be more easily achieved with only 20k bilingual sentence pairs using semi-supervised learning.
This might be due to that Russian and English are more distant to each other than German and English, thus bilingual training signal is more crucial for Russian$\leftrightarrow$English.

Note that for these two language pairs, the bilingual data for supervised learning are from many different text domains, whereas the monolingual data are from exactly the same domain of the test sets.
Even with such an advantage, the large-scale unsupervised NMT cannot compete with supervised NMT with tiny out-of-domain bilingual data.

\subsection{Monolingual Data Size}
\label{sec:exp:size}

\begin{figure}[!ht]
    \centering
    \hspace{-0.3cm}
    \includegraphics[width=0.4\textwidth]{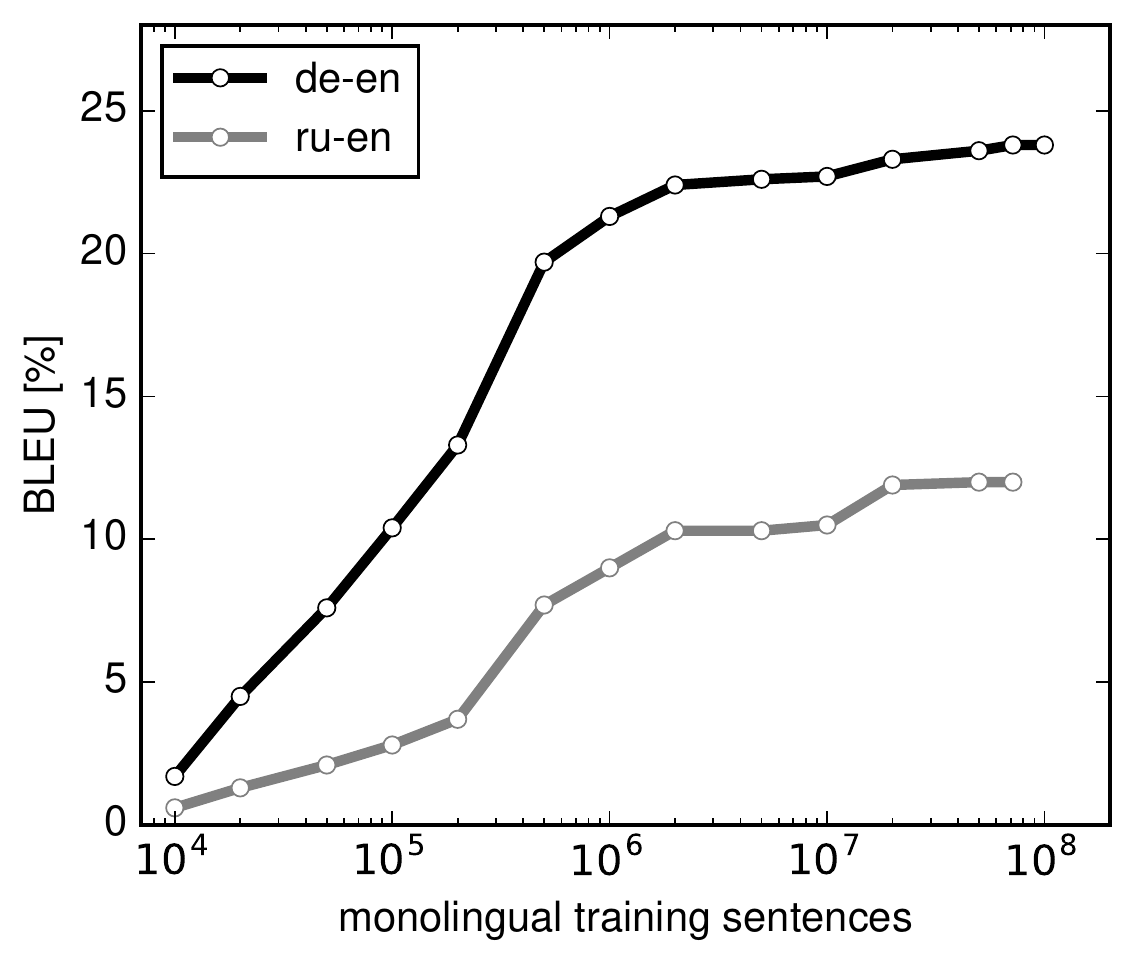}
    \caption{Unsupervised NMT performance over the size of monolingual training data, where source and target sides have the same size.}
    \label{fig:data-size-both}
    \vspace{-0.5em}
\end{figure}

In this section, we analyze how much monolingual data is necessary to make unsupervised NMT produce reasonable performance.
Figure \ref{fig:data-size-both} shows the unsupervised results with different amounts of monolingual training data.
We keep the equal size for source and target data, and the domain is also the same for both (web-crawled news).

For German$\rightarrow$English, training with only 1M sentences already gives a reasonable performance, which is only around 2\% \textsc{Bleu} behind the 100M-sentence case.
The performance starts to saturate already after 5M sentences, with only marginal improvements by using more than 20M sentences.
We observe a similar trend in Russian$\rightarrow$English.

This shows that, for the performance of unsupervised NMT, using a massive amount of monolingual data is not as important as the similarity of source and target languages.
Comparing to supervised learning (see Figure \ref{fig:unsup-sup-semisup}), the performance saturates faster when increasing the training data, given the same model size.

\subsection{Unbalanced Data Size}
\label{sec:exp:unbalanced}

What if the size of available monolingual data is largely different for source and target languages?
This is often the case for low-resource language pairs involving English, where there is plenty of data for English but not for the other side.

\begin{figure}[!ht]
    \centering
    \hspace{-0.3cm}
    \includegraphics[width=0.4\textwidth]{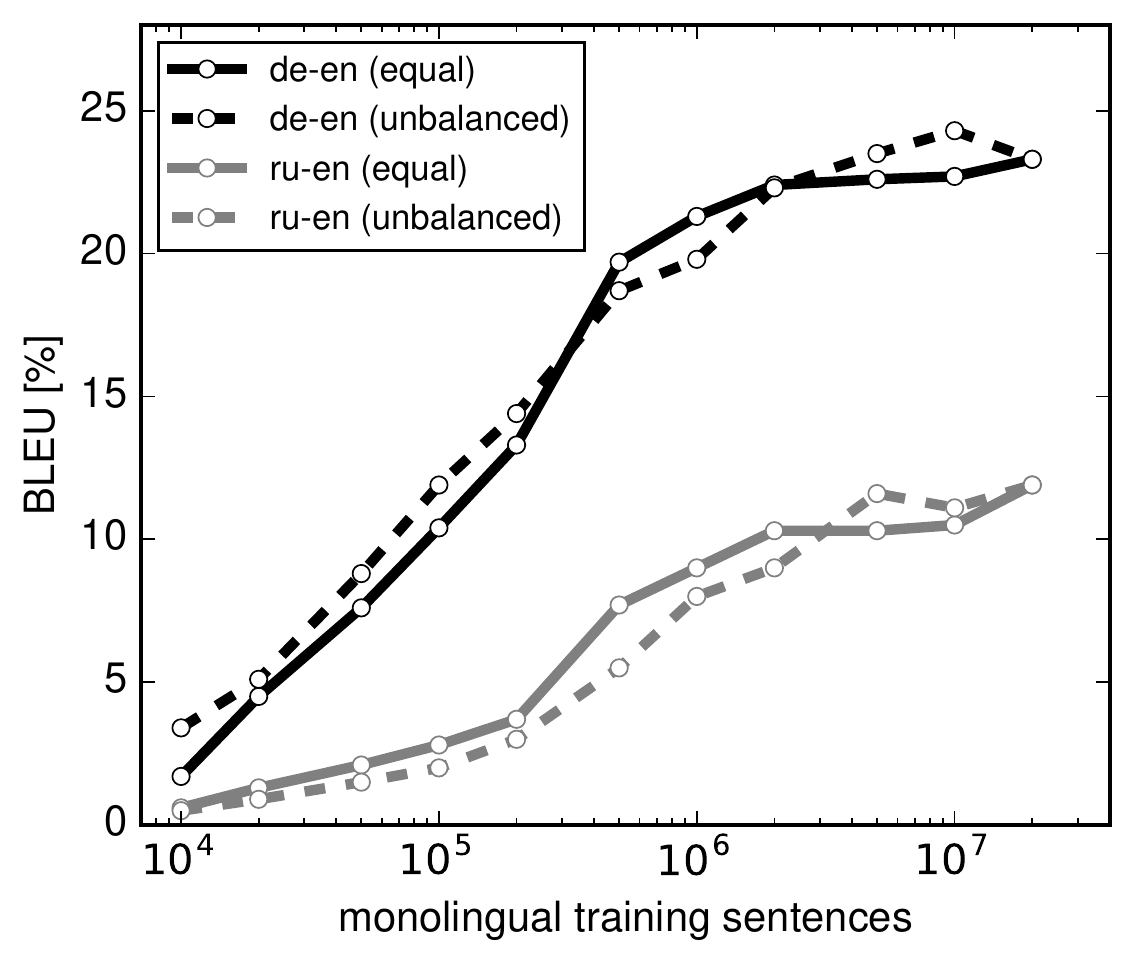}
    \caption{Unsupervised NMT performance over source training data size, where the target training data is fixed to 20M sentences (dashed line). Solid line is the case where the target data has the same number of sentences as the source side.}
    \label{fig:data-size-unbalanced}
    \vspace{-0.4em}
\end{figure}

Our experiments so far intentionally use the same number of sentences for both sides.
In Figure \ref{fig:data-size-unbalanced}, we reduced the source data gradually while keeping the large target data fixed.
To counteract the data imbalance, we oversampled the smaller side to make the ratio of source-target 1:1 for BPE learning and mini-batch construction \cite{conneau2019crosslingual}.
We compare such unbalanced data settings to the previous equal-sized source/target settings.

Interestingly, when we decrease the target data accordingly (balanced, solid line), the performance is similar or sometimes better than using the full target data (unbalanced, dashed line).
This means that it is not beneficial to use oversized data on one side in unsupervised NMT training.

If the data is severely unbalanced, the distribution of the smaller side should be much sparser than that of the larger side.
The network tries to generalize more on the smaller data, reserving the model capacity for smoothing \cite{olson2018modern}.
Thus it learns to represent a very different distribution of each side, which is challenging in a shared model (Section \ref{sec:unmt:bidir}).
This could be the reason for no merit in using larger data on one side.

\subsection{Domain Similarity}
\label{sec:exp:domain}

In high-resource language pairs, it is feasible to collect monolingual data of the same domain on both source and target languages.
However, for low-resource language pairs, it is difficult to match the data domain of both sides on a large scale.
For example, our monolingual data for Kazakh is mostly from Wikipedia and Common Crawl, while the English data is solely from News Crawl.
In this section, we study how the domain similarity of monolingual data on the two sides affects the performance of unsupervised NMT.

In Table \ref{tab:domain}, we artificially change the domain of the source side to politics (UN Corpus\footnote{https://conferences.unite.un.org/uncorpus}) or random (Common Crawl), while keeping the target domain fixed to newswire (News Crawl).
The results show that the domain matching is critical for unsupervised NMT.
For instance, although German and English are very similar languages, we see the performance of German$\leftrightarrow$English deteriorate down to -11.8\% \textsc{Bleu} by the domain mismatch.

\begin{table}[!ht]
  \centering
  \fontsize{9.5}{11}\selectfont 
  \setlength\tabcolsep{4pt}
  \begin{tabular}{ccccccc}
    \toprule
    Domain & Domain & & \multicolumn{4}{c}{\textsc{Bleu} [\%]}\\
    (\textbf{en}) & (\textbf{de}/\textbf{ru}) & & {\bf de-en} & {\bf en-de} & {\bf ru-en} & {\bf en-ru}\\
    \midrule
    \multirow{3}{*}{Newswire} & Newswire & & 23.3 & 19.9 & 11.9 & 9.3\\
    & Politics & & 11.5 & 12.2 & 2.3 & 2.5\\
    & Random & & 18.4 & 16.4 & 6.9 & 6.1\\
    \bottomrule
  \end{tabular}
\caption{Unsupervised NMT performance where source and target training data are from different domains. The data size on both sides is the same (20M sentences).}
\label{tab:domain}
\vspace{-0.2em}
\end{table}

Table \ref{tab:domain-zh} shows a more delicate case where we keep the same domain for both sides (newswire) but change the providers and years of the news articles.
Our monolingual data for Chinese (Table \ref{tab:corpus}) consist mainly of News Crawl (from years 2008-2018) and Gigaword 4th edition (from years 1995-2008).
We split out the News Crawl part (1.7M sentences) and trained an unsupervised NMT model with the same amount of English monolingual data (from News Crawl 2014-2017).
Surprisingly, this experiment yields much better results than using all available data.
Even if the size is small, the source and target data are collected in the same way (web-crawling) from similar years (2010s), which seems to be crucial for unsupervised NMT to work.

\begin{table}[!ht]
  \centering
  \fontsize{9.5}{11}\selectfont 
  \setlength\tabcolsep{4pt}
  \begin{tabular}{cccccc}
    \toprule
    Years & Years & \#sents & & \multicolumn{2}{c}{\textsc{Bleu} [\%]}\\
    (\textbf{en}) & (\textbf{zh}) & (\textbf{en}/\textbf{zh}) & & {\bf zh-en} & {\bf en-zh}\\
    \midrule
    \multirow{2}{*}{2014-2017} & 2008-2018 & \enspace1.7M & & 5.4 & 15.1\\
    & 1995-2008 & 28.6M & & 1.5 & 1.9\\
    \bottomrule
  \end{tabular}
\caption{Unsupervised NMT performance where source and target training data are from the same domain (newswire) but different years.}
\label{tab:domain-zh}
\vspace{-1em}
\end{table}

On the other hand, when using the Gigaword part (28.6M sentences) on Chinese, unsupervised learning again does not function properly.
Now the source and target text are from different decades; the distribution of topics might be different.
Also, the Gigaword corpus is from traditional newspaper agencies which can have a different tone from the online text of News Crawl.
Despite the large scale, unsupervised NMT proves to be sensitive to a subtle discrepancy of topic, style, period, etc. between source and target data.

These results agree with \newcite{sogaard2018limitations} who show that modern cross-lingual word embedding methods fail in domain mismatch scenarios.

% here for putting this in page 8
\begin{table*}[!ht]
\centering
\fontsize{7.8}{10}\selectfont 
\setlength\tabcolsep{4pt}
\begin{tabular}{ccp{4.65cm}p{4.2cm}p{3.9cm}}
\toprule
    Task & \textsc{Bleu} [\%] & Source input & System output & Reference output\\
    \midrule
    \multirow{4}{*}{\textbf{de-en}} & \multirow{2}{*}{23.8} & Seit der ersten \uline{Besichtigung} wurde die \textit{1.000 Quadratfu{\ss}} gro{\ss}e ... & Since the first \uline{Besichtigung}, the \textit{3,000 square} fueled ... & Since the first viewing, the 1,000sq ft flat has ...\\
    \cmidrule(l){2-5}
     & \multirow{2}{*}{10.4} & \textit{M\"{u}nchen} 1856: \textit{Vier} Karten, die Ihren Blick auf die \textit{Stadt} ver\"{a}ndern & \textit{Austrailia} 1856: \textit{Eight} things that can keep your way to the \textit{UK} & Munich 1856: Four maps that will change your view of the city\\
    \midrule
    \multirow{3}{*}{\textbf{ru-en}} & \multirow{3}{*}{12.0} & \foreignlanguage{russian}{В ходе \uline{первоочередных оператив-}\newline\uline{но-следственных мероприятий} ус-\newline тановлена личность роженицы} & The \foreignlanguage{russian}{\uline{первоочередных оператив-}\newline \uline{но-следственных мероприятий}} have been established by the dolphin & The identity of the mother was determined during preliminary investigative and operational measures\\
    \midrule
    \multirow{2}{*}{\textbf{zh-en}} & \multirow{2}{*}{\enspace1.5} & ... \begin{CJK*}{UTF8}{gbsn}\uline{调整 要 兼顾 生产 需要}和\uline{消费需求}。\end{CJK*} & ... \begin{CJK*}{UTF8}{gbsn}\uline{调整要兼顾生产需要}\end{CJK*} and \begin{CJK*}{UTF8}{gbsn}\uline{消费需}\newline\uline{求}.\end{CJK*} & ... adjustment must balance production needs with consumer demands.\\
\bottomrule
\end{tabular}
\caption{Problematic translation outputs from unsupervised NMT systems (\uline{input copying}, \textit{ambiguity in the same context}).}
\label{tab:examples}
\vspace{-0.8em}
\end{table*}

\subsection{Initialization vs. Translation Training}
\label{sec:exp:init-bt}

Thus far, we have seen a number of cases where unsupervised NMT breaks down.
But which part of the learning algorithm is more responsible for the performance: initialization (Section \ref{sec:unmt:init}) or translation training (Section \ref{sec:unmt:bt} and \ref{sec:unmt:dae})?

\begin{figure}[!ht]
    \centering
    \hspace{-0.3cm}
    \includegraphics[width=0.4\textwidth]{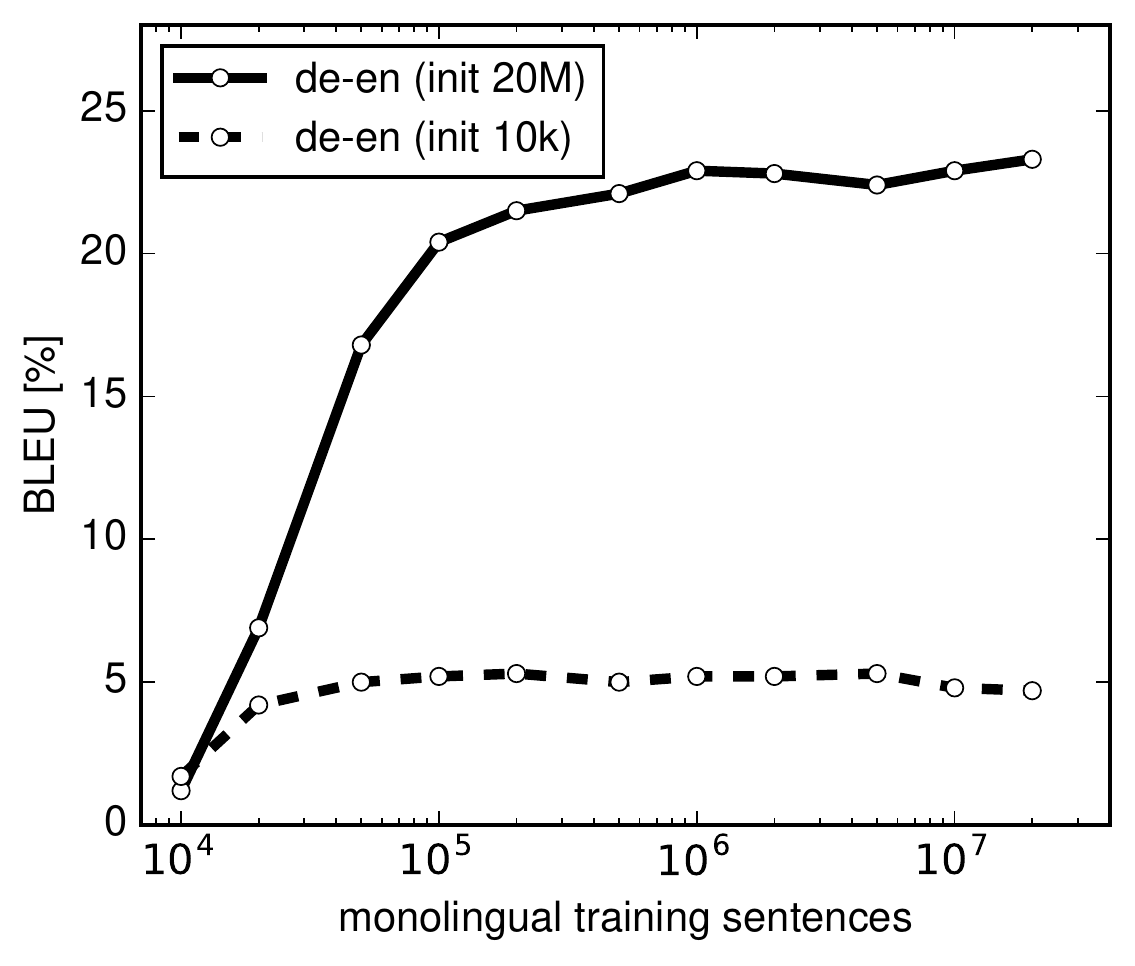}\vspace{-0.3em}
    \caption{Unsupervised NMT performance over the training data size for translation training, where the pre-training data for initialization is fixed (10k or 20M sentences).}
    \label{fig:data-size-bt}
    \vspace{-0.6em}
\end{figure}

In Figure \ref{fig:data-size-bt}, we control the level of each of the two training stages and analyze its impact on the final performance.
We pre-trained two cross-lingual LMs as initializations of different quality: bad (using 10k sentences) and good (using 20M sentences).
For each initial point, we continued the translation training with different amounts of data from 10k to 20M sentences.

From the bad initialization, unsupervised learning cannot build a reasonable NMT model, no matter how much data is used in translation training.
When the initial model is strong, it is possible to reach 20\% \textsc{Bleu} by translation training with only 100k sentences.
Using 1M sentences in translation training, the performance is already comparable to its best.
Once the model is pre-trained well for cross-lingual representations, fine-tuning the translation-specific components seems manageable with relatively small data.

\begin{figure}[!ht]
    \centering
    \hspace{-0.3cm}
    \includegraphics[width=0.4\textwidth]{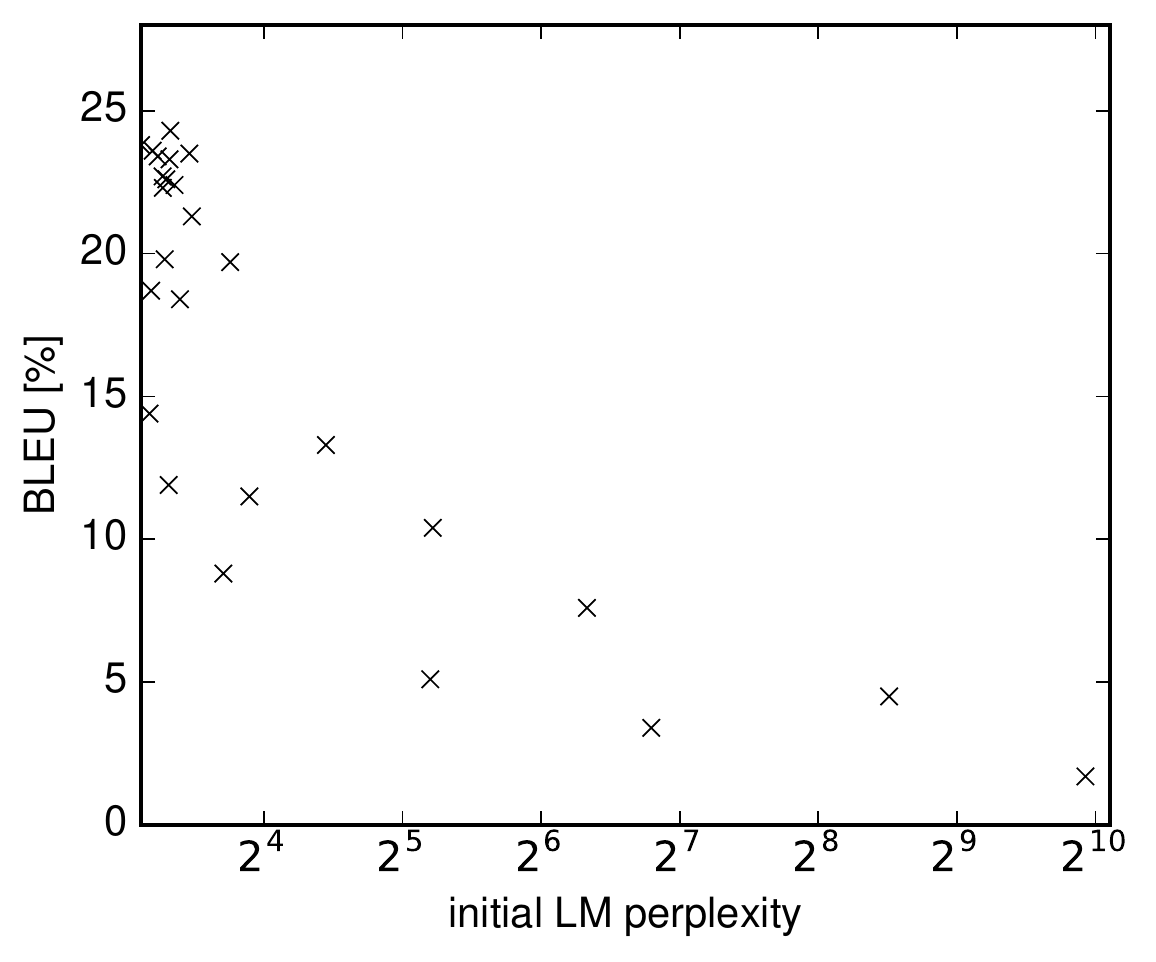}
    \caption{Unsupervised NMT performance over the validation perplexity of the initial cross-lingual LM (\textbf{de-en}).}
    \label{fig:ppl-bleu}
    \vspace{-1em}
\end{figure}

This demonstrates the importance of initialization over translation training in the current unsupervised NMT.
Translation training relies solely on model-generated inputs, i.e. back-translations, which do not reflect the true distribution of the input language when generated with a poor initial model.
On Figure \ref{fig:ppl-bleu}, we plot all German$\rightarrow$English unsupervised results we conducted up to the previous section.
It shows that the final performance generally correlates with the initialization quality.

\subsection{Qualitative Examples}
\label{sec:exp:examples}

In this section, we analyze translation outputs of unsupervised systems to find out why they record such low \textsc{Bleu} scores.
Do unsupervised systems have particular problems in the outputs other than limited adequacy/fluency?

Table \ref{tab:examples} shows translation examples from the unsupervised systems.
The first notable problem is copying input words to the output.
This happens when the encoder has poor cross-linguality, i.e. does not concurrently model two languages well in a shared space.
The decoder then can easily detect the input language by reading the encoder and may emit output words in the same language.

A good cross-lingual encoder should not give away information on the input language to the decoder.
The decoder must instead rely on the ouptut language embeddings or an indicator token (e.g. \texttt{<2en>}) to determine the language of output tokens.
As a simple remedy, we removed the language embeddings from the encoder and obtained consistent improvements, e.g. from 4.3\% to 11.9\% \textsc{Bleu} in Russian$\rightarrow$English.
However, the problem still remains partly even in our best-performing unsupervised system (the first example).

The copying occurs more often in inferior systems (the last example), where the poor initial cross-lingual LM is the main reason for the worse performance (Section \ref{sec:exp:init-bt}).
Note that the auto-encoding (Section \ref{sec:unmt:dae}) also encourages the model to generate outputs in the input language.
We provide more in-depth insights on the copying phenomenon in the appendix (Section \ref{app:copying}).

Another problem is that the model cannot distinguish words that appear in the same context.
In the second example, the model knows that \textit{Vier} in German (\textit{Four} in English) is a number, but it generates a wrong number in English (\textit{Eight}).
The initial LM is trained to predict either \textit{Four} or \textit{Eight} given the same surrounding words (e.g. 1856, things) and has no clue to map \textit{Four} to \textit{Vier}.

The model cannot learn these mappings by itself with back-translations.
This problem can be partly solved by subword modeling \cite{bojanowski2017enriching} or orthographic features \cite{riley2018orthographic,artetxe2019effective}, which are however not effective for language pairs with disjoint alphabets.

\section{Conclusion and Outlook}
\label{sec:conc}

In this paper, we examine the state-of-the-art unsupervised NMT in a wide range of tasks and data settings.
We find that the performance of unsupervised NMT is seriously affected by these factors:
\begin{itemize}\itemsep0em
    \item Linguistic similarity of source and target languages
    \item Domain similarity of training data between source and target languages
\end{itemize}
It is very hard to fulfill these in low-/zero-resource language pairs, which makes the current unsupervised NMT useless in practice.
We also find that the performance is not improved by using massive monolingual data on one or both sides.

In practice, a simple, non-tuned semi-supervised baseline with only less than 50k bilingual sentence pairs is sufficient to outperform our best large-scale unsupervised system.
At this moment, we cannot recommend unsupervised learning for building MT products if there are at least small bilingual data.

For the cases where there is no bilingual data available at all, we plan to systematically compare the unsupervised NMT to pivot-based methods \cite{kim2019pivot,currey2019zero} or multilingual zero-shot translation \cite{johnson2017google,aharoni2019massively}.

To make unsupervised NMT useful in the future, we suggest the following research directions:
\vspace{0.7em}

\noindent\textbf{Language-/Domain-agnostic LM}\hspace{0.5cm} We show in Section \ref{sec:exp:init-bt} that the initial cross-lingual LM actually determines the performance of unsupervised NMT.
In Section \ref{sec:exp:examples}, we argue that the poor performance is due to input copying, for which we blame a poor cross-lingual LM.
The LM pre-training must therefore handle dissimilar languages and domains equally well.
This might be done by careful data selection or better regularization methods.
\vspace{0.7em}

\noindent\textbf{Robust Translation Training}\hspace{0.5cm} On the other hand, the current unsupervised NMT lacks a mechanism to bootstrap out of a poor initialization.
Inspired by classical decipherment methods (Section \ref{sec:related}), we might devalue noisy training examples or artificially simplify the problem first.

\normalem
\bibliography{references}
\bibliographystyle{eamt20}
\vspace{-0.4em}

\appendix
\section{Input Copying}
\label{app:copying}

This supplement further investigates why the input copying (Section \ref{sec:exp:examples}) occurs in the current unsupervised NMT.
We discover its root cause in the unsupervised loss function, present possible remedies, and illustrate the relation to model cross-linguality with a toy experiment.

\subsection{Reconstruction Loss}
\label{app:copying:loss}

Training loss of the unsupervised NMT (Figure \ref{fig:bt}) is basically reconstruction of a monolingual sentence via an intermediate representation, created as the most probable output sequence of the current model's parameters.
This introduces an intrinsic divergence of the model's usage between in training (creating an intermediate sequence facilitates the reconstruction) and in testing (producing a correct translation).

Note that there are no constraints on the intermediate space in training.
This gives rise to a plethora of solutions to the loss optimization, which might be not aligned with the actual goal of translation.
In principle, a model could learn any bijective function from a monolingual corpus to a set of distinct sentences of the same size.

Here, input copying (Table \ref{tab:examples}) is a trivial action for the bidirectional model with a shared vocabulary, which is reinforced by training on copied back-translations.
It is easier than performing any kind of translation which might intrinsically remove information from the input sentence.

\subsection{Remedies}
\label{app:copying:remedies}

To avoid the copying behavior, we should constrain the search space of the intermediate hypotheses to only meaningful sequences in the desired language.
This is rather clear in unsupervised phrase-based MT \cite{lample2018phrasebased,artetxe2018unsupervised-smt}, where the search space is limited via the choice of applicable rules and a monolingual language model of the output language.

For unsupervised NMT, it is more difficult due to the sharing of model parameters and vocabularies over source and target languages (Section \ref{sec:unmt:bidir}).
A good initialization (Section \ref{sec:unmt:init}) and denoising autoencoder (Section \ref{sec:unmt:dae}) bias the model towards fluent outputs, but they do not prevent the model from emitting the input language.
The following techniques help to control the output language in unsupervised NMT:
\begin{itemize}\itemsep0em
    \item Restrict the output vocabulary to the desired language \cite{liu2020multilingual}
    \item Use language-specific decoders \cite{artetxe2018unsupervised-nmt,sen2019multilingual}
    \item Improve cross-linguality of the encoder, e.g. by adversarial training \cite{lample2018unsupervised}
\end{itemize}
Note that these remedies might also harm the overall training process in another respect, e.g. inducing less regularization.

\subsection{Toy Example: Case Conversion}
\label{app:copying:toy}

We empirically investigate the input copying problem with a simple task of case conversion.
In this task, the source and target languages consist only of 1-character words in lower- or uppercase respectively.
Without any constraints in back-translation, the unsupervised NMT may learn two optimal solutions to the reconstruction loss: 1) copy the casing (undesired) or 2) perform a translation from uppercase to lowercase and vice versa (desired).
We trained 1-layer, 64-dimension Transformer models with 100-sentence data on each side, and measure how often the model fails to converge to the desired solution.

To see the impact of cross-linguality, we compare two initializations where lower- and uppercase character embeddings are equally or separately initialized.
When they are equal, the model always found the desired solution (case conversion) in 10 out of 10 trials, whereas the separate variant only found it in 2 out of 10 trials.
On convergence, all experiments achieved zero reconstruction loss.

These results are in line with the language embedding removal (Section \ref{sec:exp:examples}); better cross-linguality guides the model to refer to the case indicator on the target side, which leads to case conversion.

\end{document}